\documentclass[conference]{IEEEtran}
\IEEEoverridecommandlockouts
\usepackage{cite}
\usepackage{amsmath,amssymb,amsfonts}
\usepackage{algorithmic}
\usepackage{graphicx}
\usepackage{textcomp}
\usepackage{xcolor}

\usepackage{lineno,hyperref}
\usepackage{graphicx}
\usepackage{caption}
\usepackage{subcaption}
\usepackage{graphicx}
\modulolinenumbers[5]

\def\BibTeX{{\rm B\kern-.05em{\sc i\kern-.025em b}\kern-.08em
    T\kern-.1667em\lower.7ex\hbox{E}\kern-.125emX}}
\begin{document}

\title{Robust Glare Detection: Review, Analysis, and Dataset Release\\}

\author{\IEEEauthorblockN{Mahdi Abolfazli Esfahani}
\IEEEauthorblockA{\textit{School of Electrical and Electronics Engineering} \\
\textit{Nanyang Technological University}\\
Singapore \\
mahdi001@e.ntu.edu.sg}
\and
\IEEEauthorblockN{Han Wang}
\IEEEauthorblockA{\textit{School of Electrical and Electronics Engineering} \\
\textit{Nanyang Technological University}\\
Singapore \\
hw@ntu.edu.sg}
}

\maketitle

\begin{abstract}
Sun Glare widely exists in the images captured by unmanned ground and aerial vehicles performing in outdoor environments. The existence of such artifacts in images will result in wrong feature extraction and failure of autonomous systems. Humans will try to adapt their view once they observe a glare (especially when driving), and this behavior is an essential requirement for the next generation of autonomous vehicles. The source of glare is not limited to the sun, and glare can be seen in the images captured during the nighttime and in indoor environments, which is due to the presence of different light sources; reflective surfaces also influence the generation of such artifacts. The glare's visual characteristics are different on images captured by various cameras and depend on several factors such as the camera's shutter speed and exposure level. Hence, it is challenging to introduce a general - robust and accurate - algorithm for glare detection that can perform well in various captured images. This research aims to introduce the first dataset for glare detection, which includes images captured by different cameras. Besides, the effect of multiple image representations and their combination in glare detection is examined using the proposed deep network architecture. The released dataset is available at https://github.com/maesfahani/glaredetection 
\end{abstract}


\section{Introduction}
Autonomous robots are growing exponentially, and they are becoming a meaningful part of people's daily life. Safe execution of such autonomous systems is a must, and researchers should pay more attention to figuring out risky situations and handling such cases properly. One of such dangerous situations is the existence of glare in the images. Such artifacts will damage the whole or part of the rich visual information. They can also cause wrong feature extraction, which results in the failure of navigation or localization (especially odometry) systems in autonomous robots. 
\begin{figure}[hbt!]
    \centering
    \includegraphics[width=0.5\textwidth]{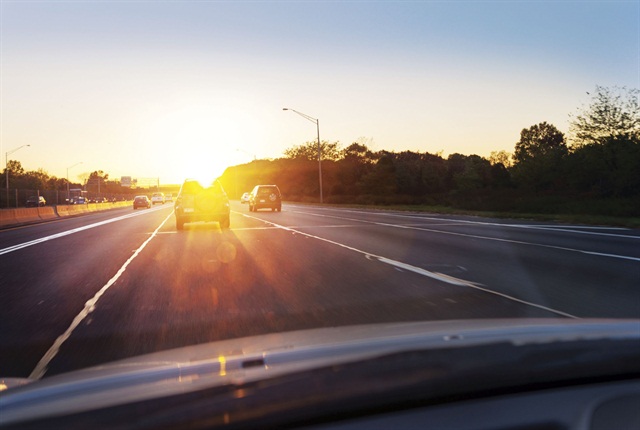}
    \caption{Sample Image Affected by Glare in the Released Dataset.}
    \label{fig:glare}
\end{figure}

Humans widely interact with sun glare while driving on the roads. Once humans see a sun glare, they immediately slow down to drive safely with missed visual information, and they turn their sight to avoid the worst effect of sun glare on visual information. This behavior is a significant behavior that needs to be modeled for full autonomy in the next generation of autonomous robots to have safe and reliable robots in our communities. It is essential to mention that the source of glare is not limited to the sun, and various sources of lights and reflective surfaces can cause such artifacts in captured images. Figure \ref{fig:glare} shows a sample glare image, and as can be seen, some circular artifacts and firm edges can be seen in the picture, which significantly affects the central core of autonomous systems.

\begin{figure*}
    \centering
    \includegraphics[width=0.7\textwidth]{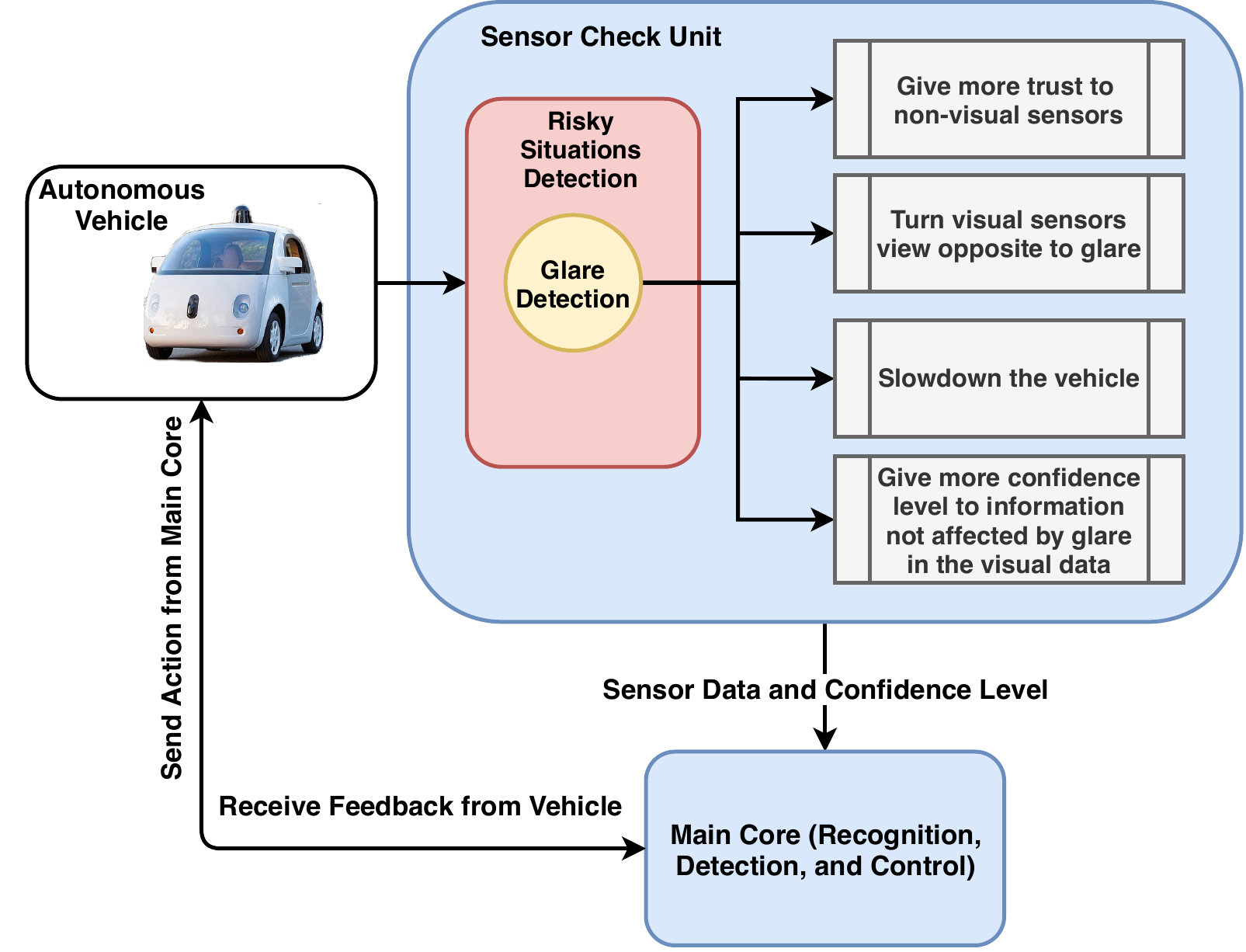}
    \caption{General Pipeline of Autonomous Vehicles. Effect of Sensor Check Unit in Detecting Risky Situations, and Necessary Tasks after Detecting Glare are Highlighted.}
    \label{fig:img1}
\end{figure*}

Autonomous robots are utilizing various types of sensors, such as Inertial Measurement Units (IMU), Light Detection and Ranging (LIDAR), and monocular cameras to solve the localization problem \cite{esfahani2019orinet,esfahani2019aboldeepio}; each of them has its own advantages and disadvantages, and researchers usually use the fusion of various sensors to make robust decisions \cite{zuo2019lic}. With the presence of glare in the visual information captured by a monocular or stereo camera, the core system should give more authority to IMU or LIDAR sensor to handle the localization and navigation problems; in this way, the effect of missed visual information on the localization module will be reduced. Besides, by extracting the glare boundaries from images, it is feasible to avoid the extraction of feature points from boundaries affected by glare, which results in the extraction of robust trackable features from visual information over time, and so a robust odometry and navigation system. Figure \ref{fig:img1} illustrates the structure of handling the risky situation, especially glare, in the next generation of autonomous vehicles.

Chahine and Pradalier \cite{chahine2018survey} have shown that one significant failure case of the odometry systems in outdoor environments is the existence of solar glares. The glares will result in wrong feature matching that causes improper motion and trajectory estimation. Besides, glares can make part of images blank, which results in a lack of points in the optimization loop for trajectory estimation. Later, Wu and Pradalier \cite{wu2019illumination} modeled sun glare as local illumination changes and reduced their effect by proposing an odometry algorithm robust to such local changes. While their method is effective, its computational cost is more than others due to using doubled residual and Jacobian computational load on their proposed cost function.

Andalibi and Chandler \cite{andalibi2017automatic} have worked on sun glare segmentation in RGB images. They have benefited from intensity, saturation, and local contrast to define a photometric map and extract the sun's location on the image. Besides, they have also utilized GPS information for handling the problem better; they used azimuth and elevation in conjunction with the vehicle heading and the road slope to determine the sun's position within the frame. Their method can detect the sun in the image as the only primary source of glare, and they cannot perform well with the presence of multiple sources of light; They also cannot detect glares that are occurred from reflections as well. Other researchers in this area have tried to use deep learning and image processing to detect overexposure images \cite{jatzkowski2018deep} and glare in fish-eye images \cite{yahiaoui2020let}.

Various deep semantic segmentation approaches have been proposed within the last few years, and they have achieved satisfactory results in handling the segmentation problem.  Among all proposed techniques, Faster R-CNN \cite{ren2015faster}, Mask R-CNN \cite{he2017mask} and U-Net \cite{ronneberger2015u} are the most popular techniques. Mask R-CNN is more an instance segmentation module that mainly identifies different instances each pixel belongs to that.  On the other hand, U-Net assists in the segmentation of each pixel to a certain type of object. U-Net has achieved satisfactory results for pixel segmentation with high speed in recent years and is in the focus of researchers \cite{liu2019computer,azad2019bi}. U-Net consists of two main stages: 1) applying convolutional and down-sampling layers to extract feature representations over scales, and 2) applying up-convolutional and convolutional layers to up-sample the downsampled feature representations and extract the output segments.

Finding a general solution that can be directly mounted on various cameras for robust glare detection is essential. However, camera parameters such as shutter speed and exposure level significantly influence the visual characteristics of glare in the captured images, which makes finding a general glare detection system challenging. Therefore, this research aims to improve the robust detection and segmentation of glare in images captured with different cameras. Hence, a dataset is created in this paper based on images available over the internet, captured by various types of cameras. Therefore, the glare detection model should not be biased and should work with multiple cameras to get high accuracy on this dataset. In short, the main contributions of this paper can be defined as:
\begin{itemize}
    \item Releasing a dataset that contains general images with their correspondent binary mask that illustrates the presence of glare in the image, which can be used in the supervised training of a glare detection module.
    \item Modifying U-Net network architecture to propose the best glare segmentation network architecture, which extracts features from various image representations over different branches and utilizes their combinations to detect glare.
    \item Evaluating various image representations for extraction of glare segments via the modified U-Net \cite{ronneberger2015u} deep semantic segmentation pipeline. Investigating the best image representation for different tasks is an essential requirement for a robust outcome \cite{esfahani2019deepdsair}.
\end{itemize}

\section{Proposed Method}

This section is proposing an algorithm for detecting glare segments of an input image. A modified version of U-Net network architecture is presented and utilized, and various image representations are examined on the proposed network architecture to investigate the best image representations that can assist in handling the task mentioned above. In the next subsections, the utilized image representations are reviewed, and afterward, the proposed network architecture is discussed. 

\subsection{Image Representations}
\begin{figure*}
     \centering
     \begin{subfigure}[b]{0.3\textwidth}
         \centering
         \includegraphics[width=\textwidth]{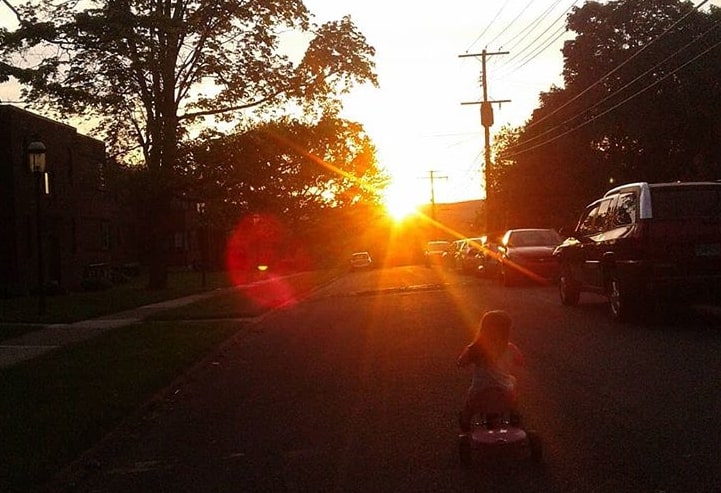}
         \caption{RGB Image}
         \label{fig:rgb}
     \end{subfigure}
     \hfill
     \begin{subfigure}[b]{0.3\textwidth}
         \includegraphics[width=\textwidth]{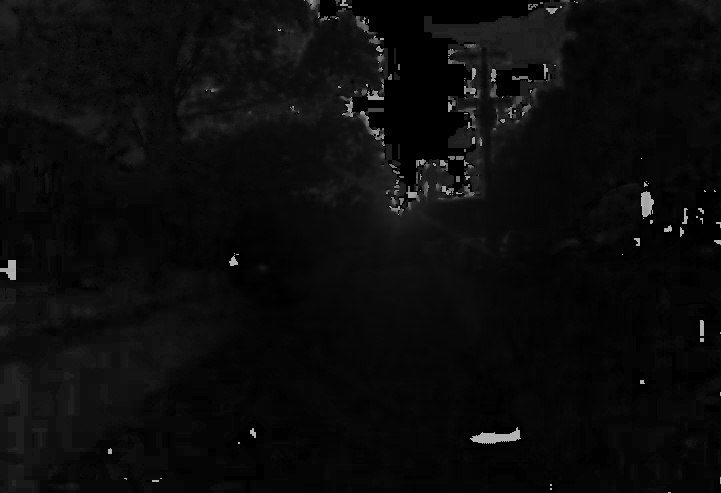}
         \caption{Hue Image}
         \label{fig:hue}
     \end{subfigure}
     \hfill
     \begin{subfigure}[b]{0.3\textwidth}
         \includegraphics[width=\textwidth]{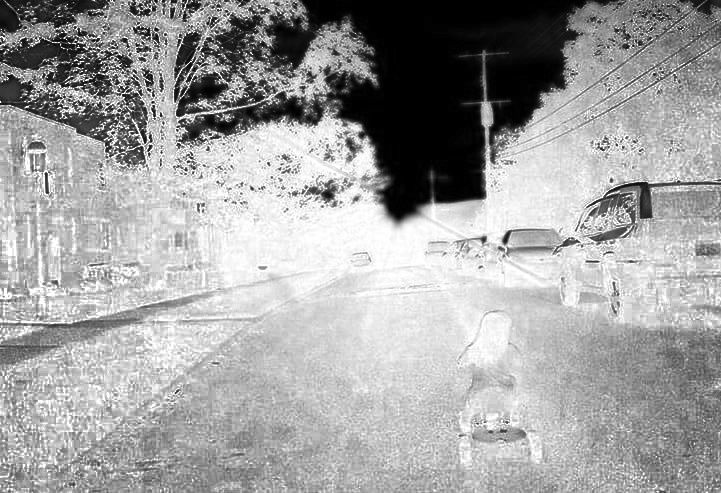}
         \caption{Saturation Image}
         \label{fig:saturation}
     \end{subfigure}
     \hfill
          \begin{subfigure}[b]{0.3\textwidth}
         \includegraphics[width=\textwidth]{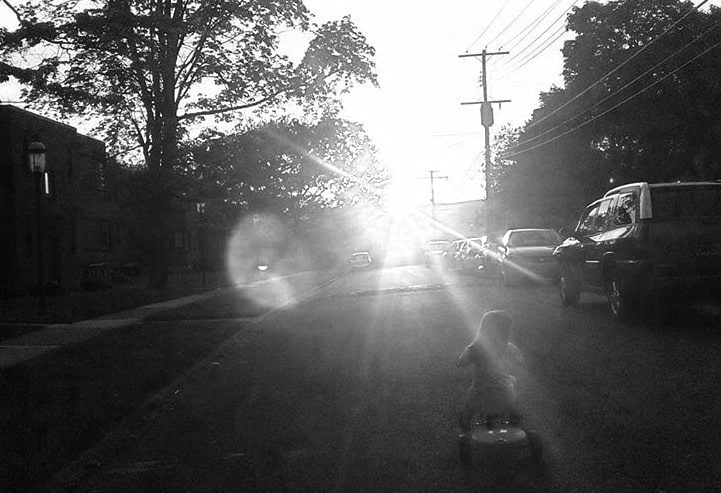}
         \caption{Intensity Image}
         \label{fig:five over x}
     \end{subfigure}
     \hfill
               \begin{subfigure}[b]{0.3\textwidth}
         \includegraphics[width=\textwidth]{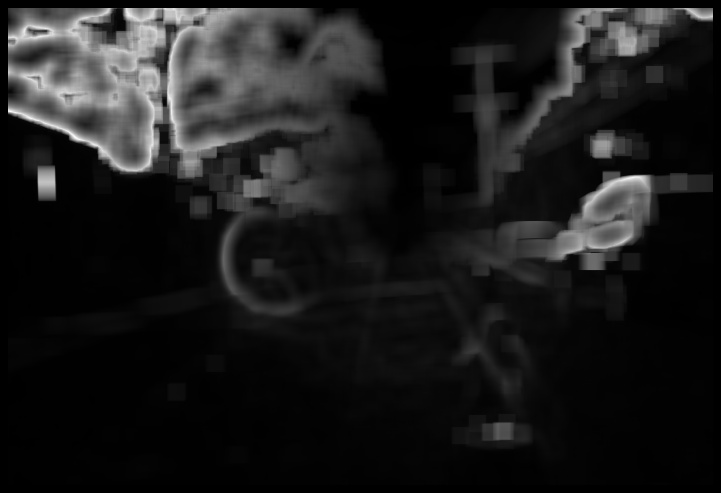}
         \caption{$I_C$}
         \label{fig:five over x}
     \end{subfigure}
      \hfill
          \begin{subfigure}[b]{0.3\textwidth}
         \includegraphics[width=\textwidth]{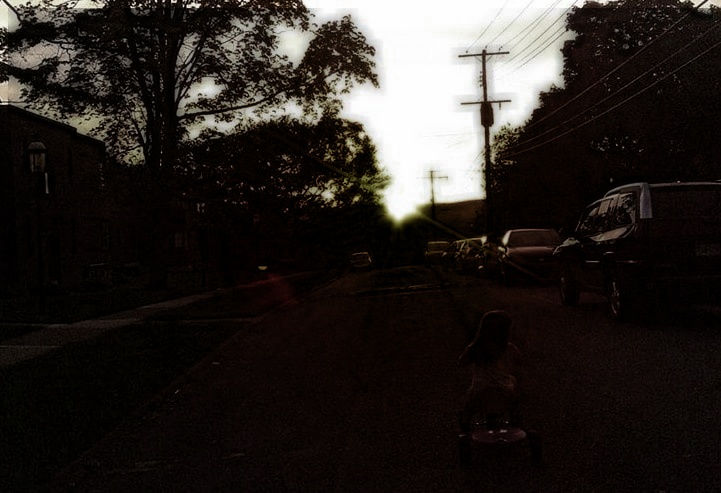}
         \caption{$I_G$}
         \label{fig:five over x}
     \end{subfigure}     
             \caption{A sample image in the dataset in the RGB color space, and it's correspondent HSV maps (b,c,d), and $I_C$ and $I_G$.}
        \label{fig:colors}
\end{figure*}

In this section, various image representations that are utilized to develop a glare segmentation module are reviewed. While the RGB color space is the most utilized color space in various computer vision applications, it cannot distinguish the Hue, Saturation, and Intensity (Value) well. Low saturation and intensity will result in a black image, while a low saturation and high intensity will result in a white image. Hence, as glares are mostly white in the captured images, the expectation is that the saturation gets low and intensity gets high in the glare segments. As such, HSV color space has rich information and is considered as one of the color spaces to study and extract glare. Hence, gathered images in the RGB color space ($I_{RGB}$) and their HSV representations ($I_{HSV}$) are the main considered images.  Figure \ref{fig:colors} is illustrating a sample image in RGB and HSV color spaces.

As claimed in \cite{andalibi2017automatic}, low luminance contrast regions have a high potential of being related to glare. Hence, the contrast map for the input image $I_{HSV}$ is computed as
\begin{equation}
    I_C(x,y) = \frac{\sqrt{\frac{1}{(N \times M) - 1}\sum_{x',y'}{[L(x',y') - \bar{L}(x,y)]^2}}}{\max(10,\bar{L}(x,y))}
\end{equation}
where
\begin{equation}
    L(x,y)~=~[0.02874 \times I_V(x,y)]^{2.2}
\end{equation}
and encounters the luminance based on the $I_V \in I_{HSV}$ (Intensity in the HSV color space). $N$ and $M$ are defining the bounding box width and height that surrounds the center pixel $(x,y)$ ($N = M = 17$ is considered in this research), and $(x',y')$ are pixels in that bounding box. The bounding box shifts on the image to get the contrast value for pixels and construct the contrast map $I_C$. Besides, the $\bar{L}$ demonstrates the average of luminance in the bounding box. The upper bound of 10 is considered for luminance to avoid disproportionately large contrasts in the division. It is important to mention that to speed up the generation of contrast map $I_C$; the bounding box can be shifted by $k$ pixels and interpolate pixels in between.

Since intensity gets high for glare, and saturation and contrast gets low, it is possible to aggregate these conditions and generate the photometric map $I_G$, which can be defined as
\begin{equation}
\label{eq:ig}
    I_G(x,y) = \textit{rescaled}(I_{RGB}(x,y) \times (1-I_S(x,y)) \times (1-I_C(x,y)))
\end{equation}
where $\textit{rescaled}$ function normalizes the output in the range of 0 and 1, and $I_S \in I_{HSV}$ (Saturation). These image representations significantly impact glare segmentation, so their influence on glare segmentation is reviewed on the proposed network architecture. Figure \ref{fig:colors} illustrates the image representations generated for the RGB image.

\subsection{Network Architecture}
\begin{figure*}[hbt!]
    \centering
    \includegraphics[width=0.8\textwidth]{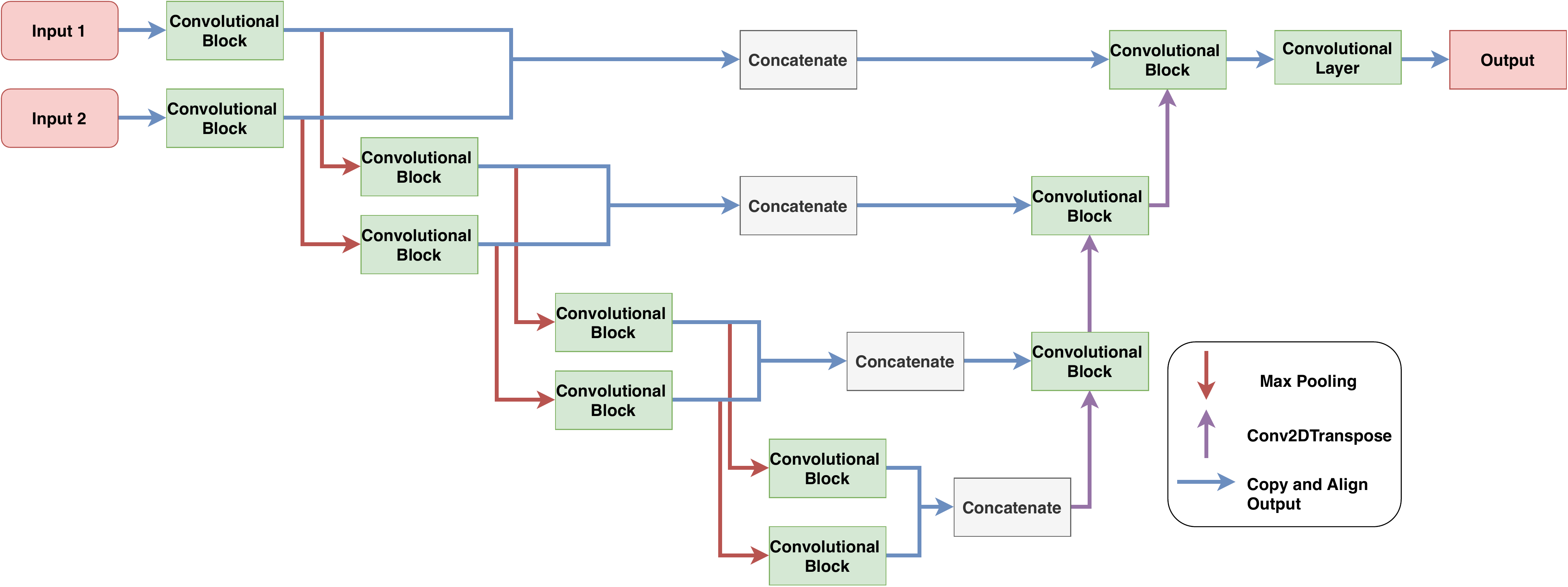}
    \caption{Modified U-Net Network Architecture.}
    \label{fig:network}
\end{figure*}

Since different image representations have different characteristics and different information needs to be extracted from each of them, a multi-branch network architecture is proposed in this subsection based on the U-Net architecture. The inputs are passed over different input branches, and the convolutional blocks apply convolutional layers to learn effective features from different inputs individually. Afterward, a max-pooling layer is applied to downsample the outcome feature representation maps. After each of the downsampling steps, there exists another set of convolutional blocks. This procedure happens until reaching a certain point for the max-pooling layers (which could be the point that downsampling is not effective anymore or not possible), and it assists in learning higher-level features over scales. Afterward, the feature representations need to be upsampled to produce the final segmentation map. The output features for different input image representations are concatenated after the last max-pooling layer's convolutional block. Then, the feature is upsampled via the convolutional transpose layer \cite{dumoulin2016guide}, and it is concatenated with the features at a similar scale in the downsampling procedure; it allows the network to do not miss any information and benefit from extracted features over scales. The convolutional layer is then applied to merge information and extract the required output in the upsampling procedure. The proposed network architecture is visualized in Figure \ref{fig:network} when two representations are input.

It is important to mention that the network is trained in a supervised manner by utilizing the weighted cross-entropy loss \cite{ronneberger2015u} considering the ground truth glare segments and the network output. It is also important to mention that, Otsu method \cite{otsu1979threshold}, which searches for the threshold that minimizes the intra-class variance among glare and non-glare class, is applied on the output of the network architecture to extract the best threshold and figure out glare parts of the image.

\section{Experiments and Results}
\subsection{Data Collection}
Since there is no dataset, to the best of our knowledge at the current stage, for glare detection, a dataset is created by gathering sample images from the world wide web and labeling the glaring part of input images. In this manner, the combination of "glare" and "sun glare" words are utilized while searching online for images. The labeling team then labels the gathered images, and the glare segments of the images are highlighted in the ground-truth binary image. Then, the masks are validated by another group to make sure the labels are trustworthy. Two hundred images are gathered and labeled in this way\footnote{The dataset is available at https://github.com/maesfahani/glaredetection}. 

\subsection{Experimental Results}
\begin{figure}
    \centering
    \includegraphics[width=0.5\textwidth]{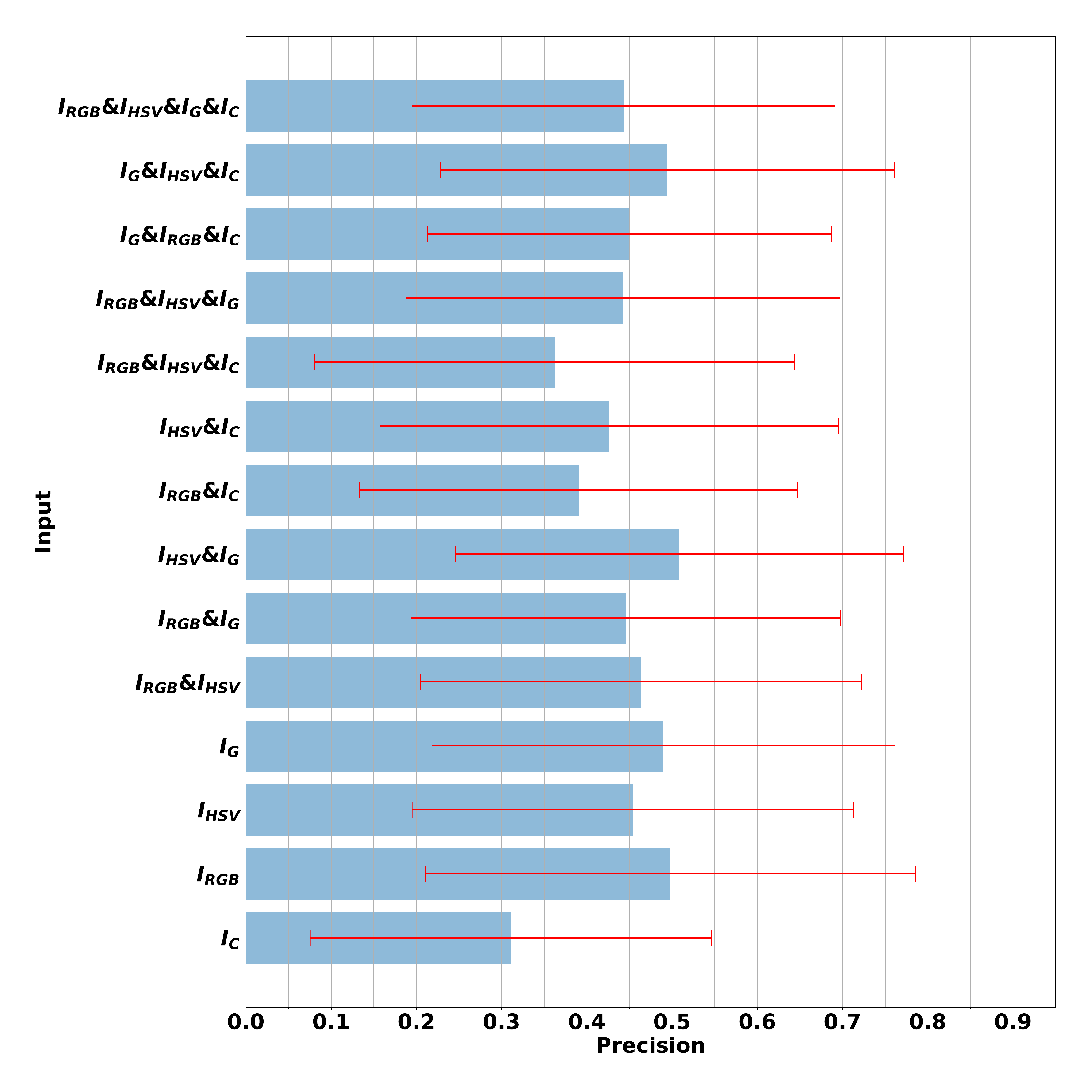}
    \caption{Comparison of Precision, with its Standard Deviation, when Different Image Representations are Utilized}
    \label{fig:per}
\end{figure}

\begin{figure}
    \centering
    \includegraphics[width=0.5\textwidth]{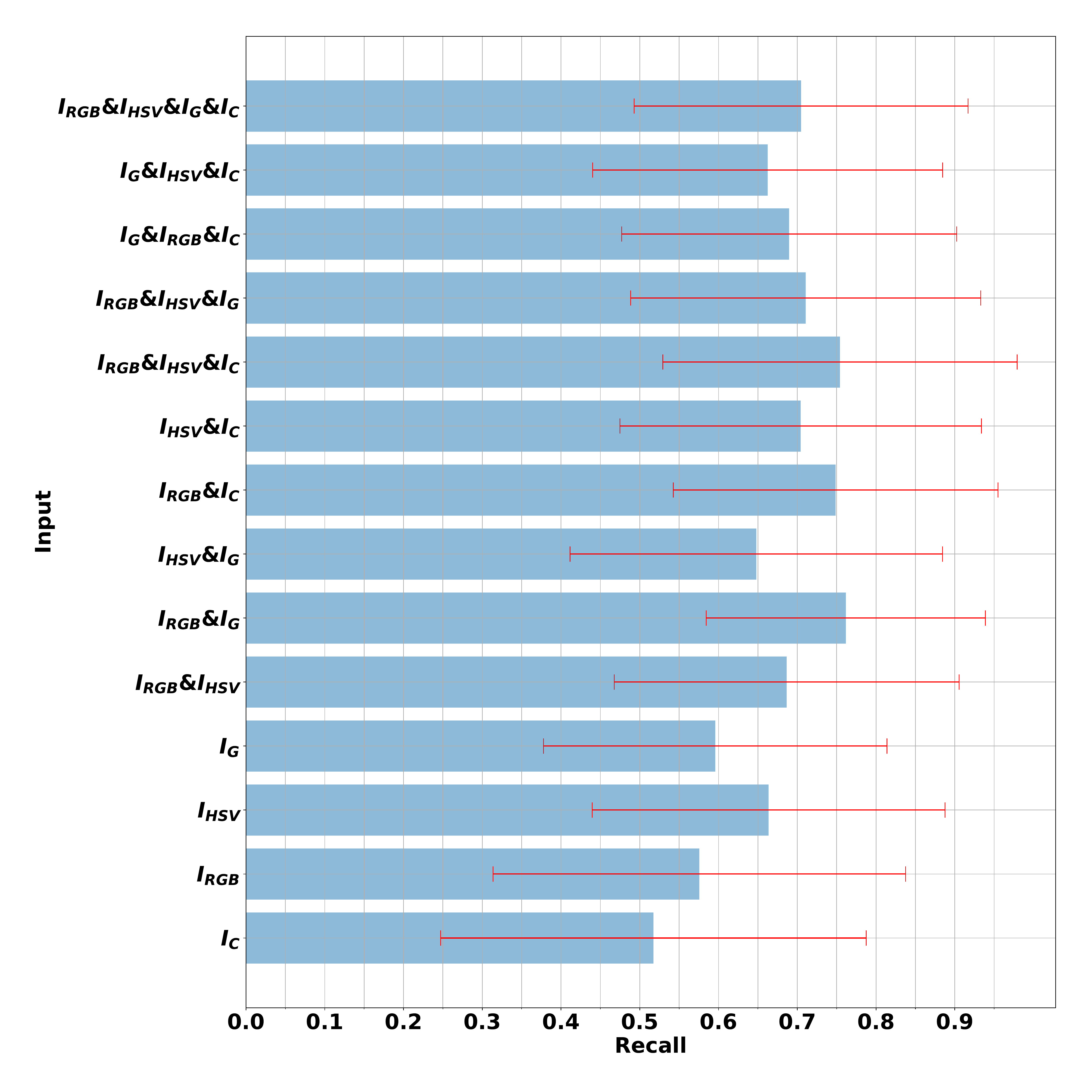}
    \caption{Comparison of Recall, with its Standard Deviation, when Different Image Representations are Utilized}
    \label{fig:rec}
\end{figure}

\begin{figure}
    \centering
    \includegraphics[width=0.5\textwidth]{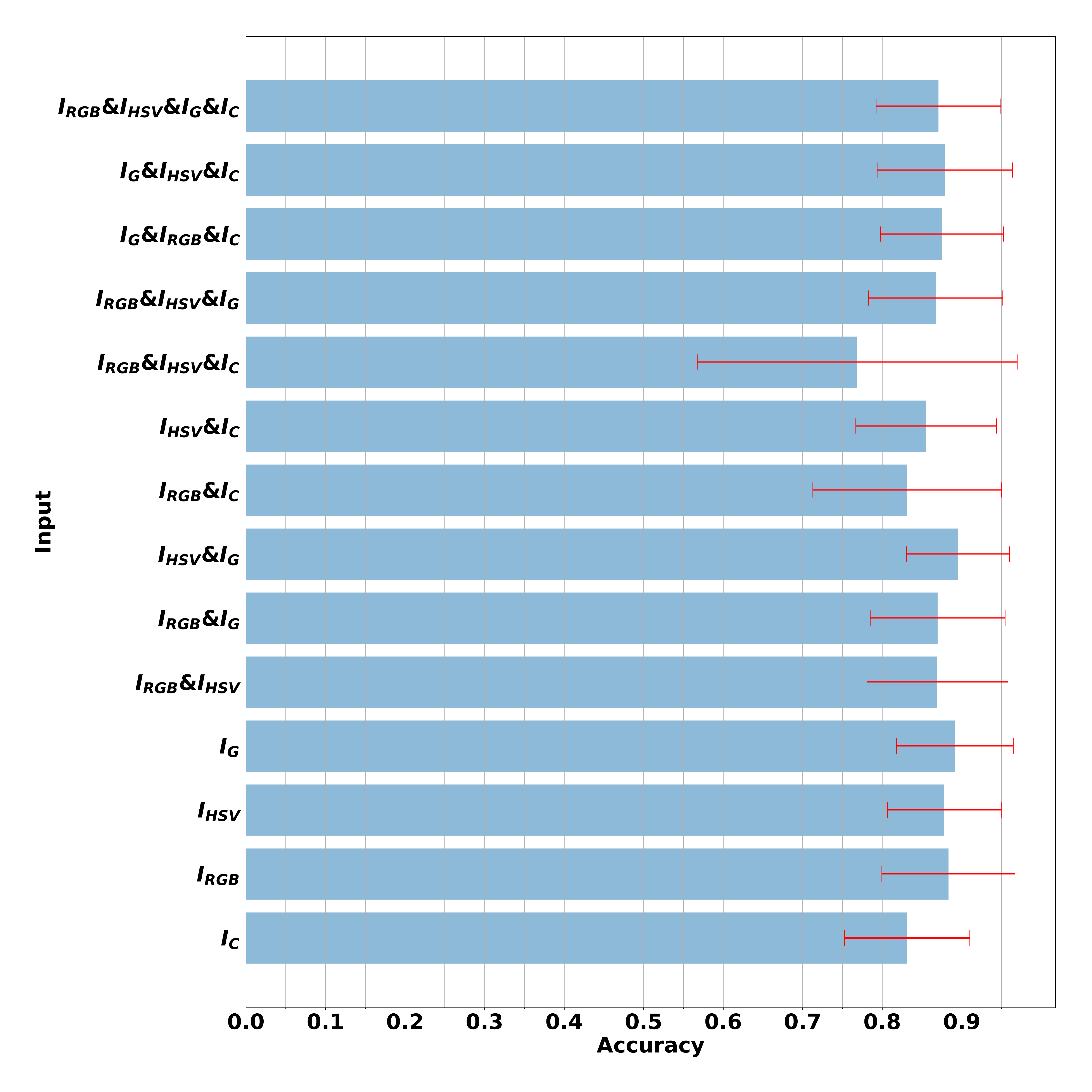}
    \caption{Comparison of Accuracy, with its Standard Deviation, when Different Image Representations are Utilized}
    \label{fig:acc}
\end{figure}

\begin{figure}
    \centering
    \includegraphics[width=0.5\textwidth]{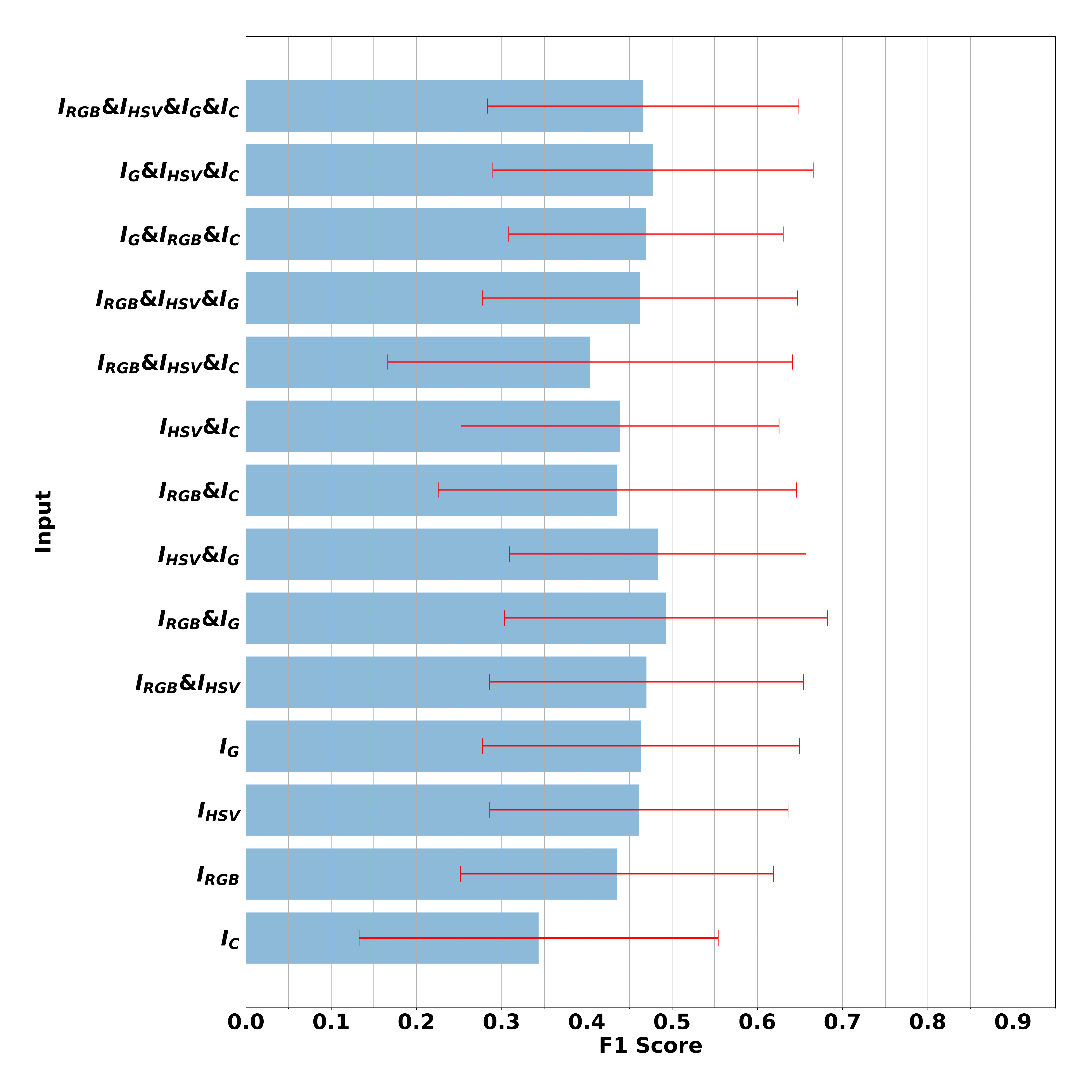}
    \caption{Comparison of F1 score, with its Standard Deviation, when Different Image Representations are Utilized}
    \label{fig:f1}
\end{figure}

The modified U-Net network architecture is trained based on the $I_{RGB}$, $I_{HSV}$, $I_{C}$, and $I_{G}$ image representations and their combination. The network is trained by utilizing an 8-fold cross-validation approach on the training data, and the averaged results can be seen in the Figures \ref{fig:per}-\ref{fig:f1}, and Table \ref{table:res} summarizes the result.

\begin{table*}[]
\fontsize{28pt}{28pt}
\selectfont
\caption{Performance Comparison when Different Image Representations are Utilized}
\resizebox{\textwidth}{!}{%
\begin{tabular}{|c|c|c|c|c|c|c|c|c|c|c|c|c|c|c|}
\cline{2-15}
\hline
\textbf{Metric}        & \textbf{$I_C$} & \textbf{$I_{RGB}$} & \textbf{$I_{HSV}$} & \textbf{$I_G$} & \textbf{$I_{RGB} \& I_{HSV}$} & \textbf{$I_{RGB} \& I_{G}$} & \textbf{$I_{G} \& I_{HSV}$} & \textbf{$I_{RGB} \& I_{C}$} & \textbf{$I_{C} \& I_{HSV}$} & \textbf{$I_{RGB} \& I_{HSV} \& I_{C}$} & \textbf{$I_{RGB} \& I_{HSV} \& I_{G}$} & \textbf{$I_{RGB} \& I_{G} \& I_{C}$} & \textbf{$I_{G} \& I_{HSV} \& I_{C}$} & \textbf{$I_{RGB} \& I_{HSV} \& I_{G} \& I_{C}$} \\ \hline
\textbf{Precision}     & 0.3107       & 0.4979           & 0.4538           & 0.4899       & 0.4634                     & 0.4458                   & \textbf{0.5083}                   & 0.3904                   & 0.4264                   & 0.3619                             & 0.4423                             & 0.4498                           & 0.4944                           & 0.4428                                   \\ \hline
\textbf{Std Precision} & 0.2355       & 0.2875           & 0.2589           & 0.2717       & 0.2586                     & 0.2520                   & 0.2628                            & 0.2569                   & 0.2691                   & 0.2814                             & 0.2545                             & 0.2371                           & 0.2664                           & 0.2481                                   \\ \hline
\textbf{Recall}        & 0.5173       & 0.5756           & 0.6636           & 0.5958       & 0.6866                     & \textbf{0.7616}          & 0.6479                            & 0.7487                   & 0.7043                   & 0.7542                             & 0.7106                             & 0.6897                           & 0.6624                           & 0.7048                                   \\ \hline
\textbf{Std Recall}    & 0.2702       & 0.2620           & 0.2240           & 0.2181       & 0.2189                     & 0.1772                   & 0.2365                            & 0.2061                   & 0.2295                   & 0.2249                             & 0.2222                             & 0.2127                           & 0.2222                           & 0.2120                                   \\ \hline
\textbf{F1}            & 0.3433       & 0.4352           & 0.4610           & 0.4635       & 0.4698                     & \textbf{0.4927}          & 0.4831                            & 0.4358                   & 0.4388                   & 0.4038                             & 0.4625                             & 0.4693                           & 0.4775                           & 0.4662                                   \\ \hline
\textbf{Std F1}        & 0.2107       & 0.1838           & 0.1750           & 0.1860       & 0.1842                     & 0.1894                   & 0.1739                            & 0.2102                   & 0.1866                   & 0.2375                             & 0.1847                             & 0.1610                           & 0.1879                           & 0.1826                                   \\ \hline
\textbf{Accuracy}      & 0.8312       & 0.8831           & 0.8781           & 0.8913       & 0.8693                     & 0.8695                   & \textbf{0.8950}                   & 0.8313                   & 0.8552                   & 0.7684                             & 0.8672                             & 0.8750                           & 0.8785                           & 0.8706                                   \\ \hline
\textbf{Std Accuracy}  & 0.0788       & 0.0836           & 0.0714           & 0.0733       & 0.0887                     & 0.0848                   & 0.0647                            & 0.1186                   & 0.0885                   & 0.2011                             & 0.0843                             & 0.0771                           & 0.0852                           & 0.0784                                   \\ \hline
\end{tabular}
}

\label{table:res}
\end{table*}

As can be seen, the best recall and F1 score are achieved by considering $I_{RGB}$ and $I_G$ at the same time. However, the best precision and accuracy is obtained when $I_G$ and $I_{HSV}$ are utilized simultaneously. As can be seen, utilizing the $I_{RGB}$, $I_{HSV}$ and $I_G$ at the same time cannot be useful and will reduce the performance; it is due to the existence of unuseful information that prevents the network from extracting robust features. The precision and accuracy of $I_{HSV} \& I_G$ is high because it mainly can determine background better, while $I_{RGB} \& I_G$ results in better recall and F1 score, which makes it a better module to consider for segmentation.

Looking at each image representation's individual effect, it can be seen that $I_{RGB}$ results in more Precision and Accuracy, while the highest F1 score is for $I_G$, and $I_{HSV}$ gives the highest recall. Looking deeper, it can be seen that the $I_G$ and $I_{HSV}$ results are near the same, which can indicate that they have similar rich information that assists the network to solve the problem; it can also be seen from $I_G$ formula (Eq. \ref{eq:ig}) that $V \in HSV$ is utilized as the most influential factor of HSV color space in detecting glare; hence, the network did not improve by combining and using them simultaneously. However, the network's performance improved, in terms of F1 score, by considering $I_G \& I_{RGB}$ with more dissimilar information. In sum, the most impactful representation is $I_G \& I_{RGB}$. 

\section{Conclusion}

This paper has shown the importance of utilizing various color spaces for segmenting glare in images captured by different cameras. The best combination for glare detection is achieved by using the photometric image representation and RGB image simultaneously; it has shown that utilizing all image representation is not good, and the network has achieved its best accuracy with the presence of input representations with uncorrelated information that can assist in improving the overall accuracy. Moreover, a database for glare segmentation is introduced.

\section*{Acknowledgement}
The authors would like to thank students at Nanyang Technological University who helped in validating and generating the dataset especially Wong, Ezekiel Ngan Seng who helped on this project as his final project  \cite{wong2020mobile}.

\bibliography{mybibfile}

\end{document}